\documentclass[runningheads]{llncs}
\usepackage[T1]{fontenc}
\usepackage{graphicx}
\usepackage{color}
\usepackage{cite}
\usepackage{amsmath,amssymb,amsfonts}
\usepackage{algorithmic}
\usepackage{textcomp}

\usepackage{algorithm}
\usepackage{algorithmic}
\usepackage{multirow}
\usepackage[misc]{ifsym}
\begin{document}
\title{Coupling User Preference with External Rewards to Enable 
Driver-centered and Resource-aware EV Charging Recommendation}
\titlerunning{Coupling User Preference with External Rewards for EV Charging Rec.}
%
\author{Chengyin Li \and
	Zheng Dong \and
	Nathan Fisher \and 
	Dongxiao Zhu [\Letter]}
\authorrunning{Li et al.}
%
\institute{Department of Computer Science,  Wayne State University, Detroit 
MI 48201, USA \\
	\email{\{cyli, dong, fishern, dzhu\}@wayne.edu}}
\maketitle              
\begin{abstract}
Electric Vehicle (EV) charging recommendation that both accommodates user 
preference and adapts to the ever-changing external environment arises as a 
cost-effective strategy to alleviate the range anxiety of private EV drivers. 
Previous studies focus on centralized strategies to achieve optimized resource 
allocation, particularly useful for privacy-indifferent taxi fleets and 
fixed-route public transits. However, private EV driver seeks a more 
personalized and resource-aware charging recommendation that is tailor-made to 
accommodate the user preference (when and where to charge) yet sufficiently 
adaptive to the spatiotemporal mismatch between charging supply and demand. 
Here we propose a novel Regularized Actor-Critic (RAC) charging recommendation 
approach that would allow each EV driver to strike an optimal balance between 
the user preference (historical charging pattern) and the external reward 
(driving distance and wait time). Experimental results on two real-world 
datasets demonstrate the unique features and superior performance of our 
approach to the competing methods.

\keywords{Actor critic \and Charging recommendation \and Electric vehicle (EV) 
\and User preference \and External reward.}
\end{abstract}
\section{Introduction}

Electric Vehicles (EVs) are becoming popular due to their decreased carbon 
footprint and intelligent driving experience over conventional internal 
combustion vehicles \cite{o2018electric} in personal transportation tools. 
Meanwhile, the miles per charge of an EV is limited by its battery capacity, 
together with sparse allocations of charging stations (CSs) and excessive 
wait/charge time, which are major driving factors for the so-called range 
anxiety, especially for private EV drivers. Recently, developing intelligent 
driver-centered charging recommendation algorithms are emerging as a 
cost-effective strategy to ensure sufficient utilization of the existing 
charging infrastructure and satisfactory user experience 
\cite{wang2016electric, wang2018bcharge}. 

\begin{figure}[t]
	\centering
	\includegraphics[scale=0.85]{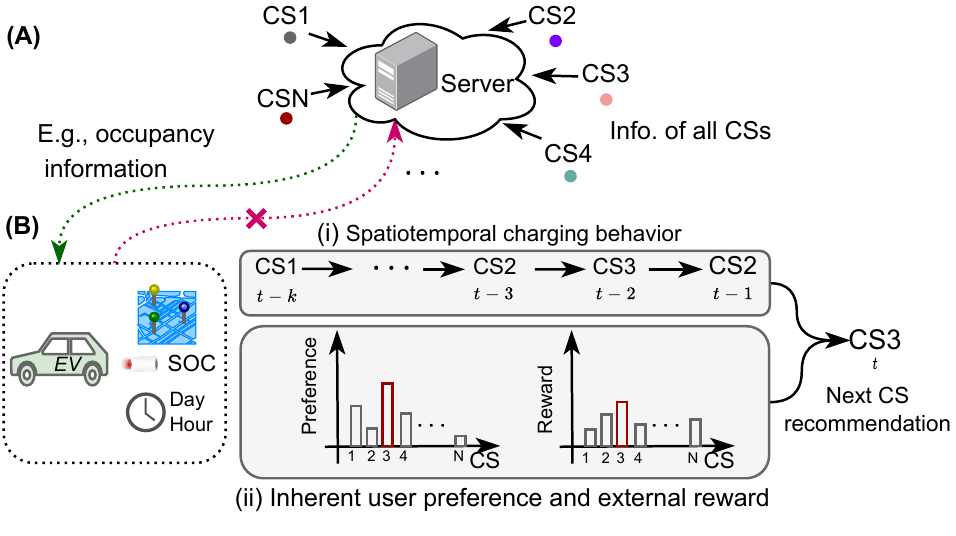}
	\caption{Driver-centered and resource-aware charging recommendation. (A) 
	Centralized charging recommendation enables optimized resource allocation, 
	where bi-directional information sharing between the sever and EVs is 
	assumed. (B) Driver-centered charging recommendation considers user 
	preference and external reward, where only monodirectional information 
	(e.g., the occupancy information of all CSs) sharing from the sever to each 
	EV is required (green dotted line). Therefore, private information of an 
	EV, like GPS location, is not uploaded to the server (pink dotted line).}
	\label{fig:general}
\end{figure}

Existing charging recommendation studies mainly focus on public EVs 
(e.g., electric taxis and buses) \cite{wang2018bcharge, guo2017recommendation}. 
With relatively fixed schedule routines, and no privacy or user preference 
consideration, the public EV charging recommendation for public transits can be 
made completely to optimize CS resource utilization. In general, these 
algorithms often leverage a global server, which monitors all the CSs in a city 
(Fig. \ref{fig:general}A). Charging recommendation can be fulfilled upon 
requests for public EVs by sending their GPS locations and state of charge 
(SOC). This kind of recommendation gives each EV an optimal driving and wait 
time before charging. Instead of using one single global server, many servers 
can be distributed across a city 
\cite{cao2018decentralized,guo2017recommendation} to reduce the recommendation 
latency for public EVs.  

Although server-centralized methods have an excellent resource-aware property 
for the availability of charging for CSs, for private EVs, they rarely 
accommodate individual user preferences of charging and even have the risk of 
private data leakage (e.g., GPS location). Thus, the centralized strategy would 
also impair the trustworthiness \cite{qiangcounterfactual2022,li2021improving, 
pan2021explaining} of the charging recommendation. A driver-centered instead of 
a server-centralized charging recommendation strategy would be preferred for a 
private EV to follow its user preference without leaking private information. 
In this situation (Fig. \ref{fig:general}B), there would be a sequence of 
on-EV charging events records (when and which CS) that reflect the personal 
preference of charging patterns for a private EV driver. To enable the 
resource-aware property for a driver-centered charging recommendation, creating 
a public platform for sharing availability of CSs is needed.   


Motivated by the success of recent research on the next POI (Point Of Interest) 
recommendation centered on each user, these studies can also be adapted to 
solve the charging recommendation problem for private EVs when viewing each CS 
as a POI. Different from collaborative filtering, based on the general 
recommendation that learns similarities between users and items 
\cite{pan2020explainable}, the following POI recommendation algorithms attempt 
to predict the most likely next POI that a user will visit based on the 
historical trajectory 
\cite{zhao2020discovering,rendle2010factorizing,zhu2017next,
 kong2018hst,zhao2019go}. Although these methods indeed 
model user preferences, they are neither resource-aware nor adapted to the 
ever-changing external environment.

As such, a desirable charging recommender for a private EV requires: (1) 
learning the user preference from its historical charging patterns for 
achieving driver-centered recommendation, and (2) having a good external reward 
(optimal driving and wait time before charging) to achieve resource-aware 
recommendation (Fig. \ref{fig:general} B). By treating the private EV 
charging recommendation as the next POI recommendation problem, maximizing 
external rewards (with a shorter time of driving and wait before charging) by 
exploring possible CSs for each recommendation, reinforcement learning can be 
utilized. To leverage user preference and external reward, we propose a novel 
charging recommendation framework, Regularized Actor-Critic (RAC), for private 
EVs. The critic is based on a resource-saving over all CSs to give a evaluation 
value over the prediction of actor representing external reward, and the actor 
is reinforced by the reward and simultaneously regularized by the driver’s user 
preference. Both actor and critic are based on deep neural networks (DNNs).

We summarize the main contributions of this work as follows: (1) we 
design and develop a novel framework RAC to give driver-centered and 
resource-aware charging recommendations on-EV recommendation; (2) RAC is 
tailor-made for each driver, allowing each to accommodate inherent user 
preference and also adapt to ever-changing external reward; and (3) we propose 
a warm-up training technique to solve the cold-start recommendation problem for 
new EV drivers.

\section{Related Work}
Next POI recommendation has attracted much attention recently in location-based 
analysis. There are two lines of POI recommendation methods: 
(1) following user preference from sequential visiting POIs 	
regularities, and (2) exploiting external incentive via maximizing the utility 	
(reward) of recommendations.   

For the first line of research, the earlier works primarily attempt to 
solve the sequential next-item recommendation problem using temporal features. 
For example, \cite{rendle2010factorizing} introduces Factorizing Personalized 
Markov Chain (FPMC) that captures sequential dependency 
between the recent and next items as well as the general taste of a user using a
combination of matrix factorization and Markov chains for next-basket 
recommendation. \cite{zhu2017next} proposes a time-related Long-Short Term 
Memory (LSTM) network to capture both long- and short-term sequential influence 
for next item recommendation. \cite{li2014modeling} attempts to 
model user' preference drift over time to achieve a better user experience in 
next item recommendation. These next-item recommendation approaches only use 
temporal features whereas next POI recommendation  would need to use both 
temporal and geospatial features.  

More recent studies of next POI recommendation not only model temporal 
relations but also consider geospatial context, such as ST-RNN 
\cite{liu2016predicting} and ATST-LSTM \cite{huang2019attention}.  
\cite{kong2018hst} proposes a hierarchical extension of LSTM to code spatial 
and temporal contexts into the LSTM for general location recommendation. 
\cite{zhao2019go} introduces a spatiotemporal gated network model where they 
leverage time gate and distance gate to control the effect of the last visited 
POI on next POI recommendation. \cite{zhao2020discovering} extends the 
gates with a power-law attention mechanism with more attention on the nearby 
POIs and explores the subsequence patterns for next POI recommendation. 
\cite{wu2019long} develops a long and short-term 
preference learning model considering sequential and context information for 
next POI recommendation. User preference-based methods can achieve 
significant performance for the following users' previous experience; however, 
they are restricted from making novel recommendations beyond users' previous 
experience.

Although few studies exploit external incentive, these methods can help explore 
new possibilities for next POI recommendation. Charging Recommendation with 
multi-agent reinforcement learning is applied for public 
EVs\cite{wang2020joint,zhang2021intelligent}, in which private information from 
each EV is inevitably required. \cite{Massimo2018HarnessingAG} 
proposes an inverse reinforcement learning method for next visit action 
recommendation by maximizing the reward that the user gains when discovering 
new, relevant, and non-popular POIs. This study utilizes the optimal POI 
selection policy (the POI visit trajectory of a similar group users) as the 
guidance. As such, it is only applicable for the centralized charging 
recommendation for privacy-indifferent public transit fleets where charging 
events are aggregated to the central server to learn the user group. However, 
this approach is not applicable to the driver-centered EV charging 
recommendation problem that we are tackling since the individual charging 
pattern is learned without data sharing across drivers. Besides the inverse 
reinforcement learning approach, \cite{Zheng2018DRNAD} introduces deep 
reinforcement learning for news recommendation, and \cite{wang2018supervised} 
proposes supervised reinforcement learning for treatment recommendation. These 
methods are also based on learning similar user groups thus not directly 
applicable to the driver-centered EV charging recommendation task, the latter 
is further subject to resource and geospatial constraints. 

Despite the existing approaches utilized spatiotemporal, social 	
network, and/or contextual information for effective next POI recommendations, 
they do not possess the desirable features for CS recommendation, which are (1) 
driver-centered: the trade-off between the driver's charging preference and the 
external reward is tuned for each driver, particularly for new drivers, and (2) 
resource-aware: there is usually capacity constraint on a CS but not on a 
social check-in POI.

\section{Problem Formulation}
Each EV driver is considered as an agent, and the trustworthy server that 
collects occupancy information of all the CSs represent the external 
ever-changing environment. We considered our charging recommendation as a 
finite-horizon MDP problem where a stochastic policy consists of a state space 
$\mathcal{S}$, an action space $\mathcal{A}$, 
and a reward function $r$: $\mathcal{S} \times \mathcal{A} \rightarrow 
\mathbb{R}$. At each time point $t$, an EV driver with the current state $s_t 
\in \mathcal{S}$, chooses an action $a_t $, i.e., the one-hot encoding of a CS, 
based on a stochastic policy $\pi_\theta(a|s)$ where $\theta$ is the set of 
parameters, and receives a reward $r_t$ from the spatiotemporal environment. 
Our objective is to learn such a stochastic policy $\pi_\theta(a|s)$ 
to select an action $a_t \sim \pi_\theta(a|s)$ by maximizing the sum of 
discounted rewards (return $R$) from the time point $t$, which is defined as 
$R_t = \sum_{i=t}^{T} \gamma^{(i-t)}r(s_i,a_i)$, and simultaneously minimizing 
the difference from the EV driver's decision $\hat{a}_t$. $\gamma \in 
[0,1]$, e.g., 0.99, is a discount factor to balance the importance of immediate 
and future rewards. $T$ is the furthermost time point we use.

The charging recommendation task is a process to learn a good 
policy for next CS recommendation for an EV driver. By modeling user behaviors 
with situation awareness, two types of methods can be designed 
to learn the policy: value based Reinforcement Learning (RL) to maintain a 
greedy policy, and policy gradient based RL to learn a parameterized stochastic 
policy $\pi_\theta(a|s)$ or a deterministic policy $\mu_\theta(s)$, where 
$\theta$ represents the set of parameters of the policy. For the discrete 
property of CSs, we focus on learning a personalized stochastic policy  
$\pi_\theta(a|s)$ for each EV driver using DNNs. 

\section{Our Approach}

%

\subsection{Background}

Q-learning \cite{watkins1992q} is an off-policy learning 
strategy for solving RL	problems that finds a greedy policy $\mu(s) = \arg 
\max_aQ^\mu(s,a)$, where $Q^\mu(s,a)$ is $Q$ value or action-value, and it is 
usually used for a small discrete action space. For any finite Markov decision 
process, Q-learning finds an optimal policy in the sense of maximizing the 
expected value of the total reward over any successive steps, starting from the 
current state. The value of $Q^\mu(s, a)$ can be calculated with dynamic 
programming. With the introduction of DNNs, a deep Q network (DQN) is used to 
learn such $Q$ function $Q_w(s, a)$ with parameter $w$, and DNN is incapable of 
handling a high dimension action space. During training, a replay buffer is 
introduced for sampling, and DQN asynchronously updates a target network 
$Q_w^{tar}(s, a)$ to minimize the expectation of square loss.        

Policy gradient \cite{sutton1999policy} is another approach to 
solve RL problems and can be employed to handle continuous or high-dimensional 
discrete actions, and it targets modeling and optimizing the policy directly. 
The policy is usually modeled with a parameterized function respect to 
$\theta$, $\pi(s,a)$. The value of the reward (objective) function depends on 
this policy and then various algorithms can be applied to optimize $\theta$ for 
the best reward. To learn the parameter $\theta$ of $\pi_\theta(a|s)$, we 
maximize the expectation of state-value function $V^{\pi_\theta}(s)=\sum_{a} 
\pi_\theta(a|s)Q^{\pi_\theta}(s,a)$, where $Q^{\pi_\theta}(s,a)$ is the 
state-value function. Then we need to maximize $J(\pi_\theta)=E_{s\sim 
	\rho^{\pi_\theta}}[V^{\pi_\theta}(s_1)]$, where $\rho^{\pi_\theta}$ 
represents 
the discounted state distribution. Policy gradient learns the parameter 
$\theta$ by the gradient $\nabla_\theta J(\pi_\theta)$, which is calculated 
with the policy gradient theorem: $\nabla_\theta J(\pi_\theta) = 
\mathbb{E}_{s\sim \rho^{\pi_\theta}, a}[\nabla_\theta \log 
{\pi_\theta}(a|s)Q^{\pi_\theta}(s,a)]$. These calculations are guaranteed by 
the policy gradient theorem.

Actor-critic \cite{Schmitt2020OffPolicyAW} method combines 
the advantages of Q-learning and policy gradient to accelerate and stabilize 
the learning process in solving RL problems. It has two components: a) an 
actor to learn the parameter $\theta$ of $\pi_\theta$ in the direction of the 
gradient $\nabla J(\pi_\theta)$ to maximize $J(\pi_\theta)$, and b) a critic 
to estimate the parameter $w$ in an action-value function $Q_w(s,a)$. 

In this paper, we use an off-policy actor-critic \cite{Schmitt2020OffPolicyAW, 
wang2018supervised}, where the actor updates the policy 
weights. The critic learns an off-policy estimate of the value function for the 
current actor policy, different from the (fixed) behavior policy. The actor 
then uses this estimate to update the policy. Actor-critic methods consist of 
two models, which may optionally share parameters. Critic updates the value 
function parameters $w$ for state action-value $Q_w(s,a)$. Actor updates the 
policy parameters $\theta$ for $\pi_\theta(a|s)$, in the direction suggested by 
the critic.  $\pi_\theta(a|s)$ is obtained by averaging the state distribution 
of behavior policy $\beta(a|s)$. $\beta(a|s)$ for collecting samples is a known 
policy (predefined just like a hyperparameter). The objective function sums up 
the reward over the state distribution defined by this behavior policy: 
$J(\pi_\theta)=E_{s\sim \rho^{\pi_\beta}}[Q^\pi(s,a)\pi_{\theta}(a|s)]$, where 
$\pi_\beta$ is the stationary distribution of the behavior policy $\beta(a|s)$; 
and $Q^\pi$ is the action-value function estimated with regard to the target 
policy $\pi$. 

\begin{figure}[t]
	\centering
	
	\includegraphics[width=\textwidth]{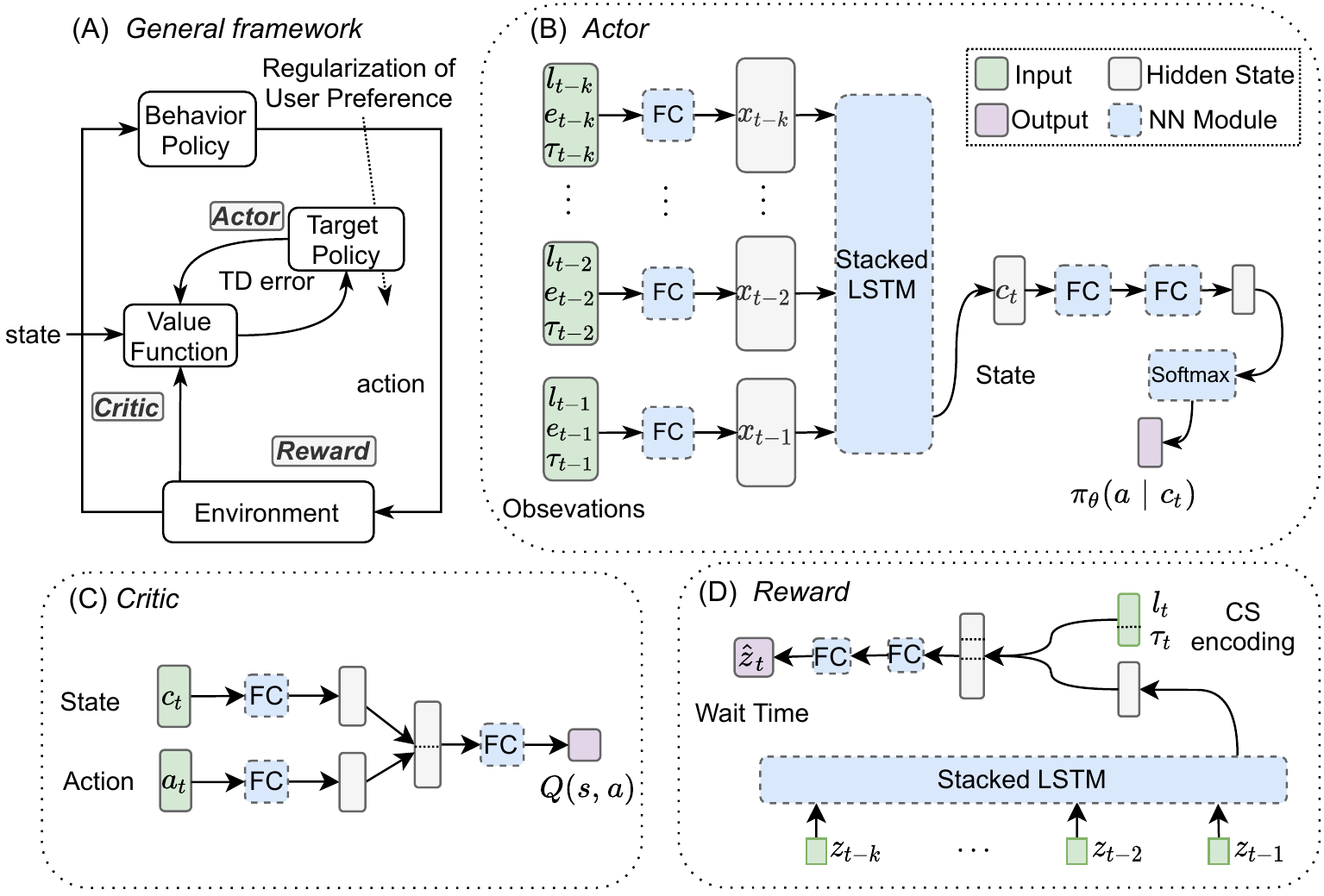}
	\caption{Our regularized actor-critic architecture. 
		(A) A general framework. (B) Actor network decides which action (CS) to 
		take (charge). (C) Critic network tells the actor how good the action 
		is and how it should be adjusted. (D) Reward network estimates wait 
		time estimation at 
		each CS.}
	\label{fig:Neural_Networks}
\end{figure}

\subsection{The Regularized Actor-Critic (RAC) Method}
To find an optimal policy for the MDP problem also with following user 
preference, we use the regularized RL method, specifically with a 
regularized actor-critic model \cite{wang2018supervised}, which combines the 
advantages of Q-learning and policy gradient. Since the computation cost 
becomes intractable with many states and actions when using policy iteration 
and value iteration, we introduce a DNN-based actor-critic model to 
reduce the computation cost and stabilize the learning. While the traditional 
actor-critic model aims to maximize the reward without considering a driver's 
preference, we also use regularization to learn the user's historical charging 
behavior as a representation of user preference. Our proposed general 
regularized actor-critic framework is shown in Fig. \ref{fig:Neural_Networks}A. 

The actor network learns a policy $\pi_\theta(a|s)$ with a set of parameters 
$\theta$ to render charging recommendation for each EV driver, where the input 
is $s_t$ and the output is the probabilities of all actions in $\mathcal{A}$ of 
transitioning to a CS $a_t$. By optimizing the two learning tasks 
simultaneously, we maximize the following objective function: 
$$J(\theta)=(1-\epsilon)J_{RL}(\theta) + \epsilon(-J_{R}(\theta)), $$ where 
$\epsilon$ is tuning parameter to weigh between inherent user preference and 
external reward (return) when making recommendation. The RL objective 
$J_{RL}$ aims to maximize the expected return via learning the policy 
$\pi_\theta(a|s)$ by maximizing the state value of an action that is averaged 
over the state distribution of the CS selection for each EV driver, i.e., 
$$ J_{RL}(\theta)=\mathbb{E}_{s \sim \rho^{\pi_\theta}, a~ \pi_\theta(a|s)} 
[Q_w(s,a)].$$ The regularization objective $J_{R}$ aims to minimize the 
discrepancy between the recommended CS and preferred CS for each user via 
minimizing the difference between CS recommended by $\pi_\theta(a|s)$ and CS 
given by each EV driver's previous selection, in terms of the cross entropy 
loss, i.e., $$J_{R}(\theta)=\mathbb{E}_{s \sim \rho^{\hat{\mu}(s)}}[-\frac{1}{K} 
\sum_{k=1}^K \hat{a}_ {t,k} \log\pi_\theta^k(a|s)-(1-\hat{a}_{t,k}) 
\log(1-\pi_\theta^k(a|s))].$$ Using DNNs, $\theta$ can be learned with 
stochastic gradient decedent (SGD) algorithms. 

The critic network is jointly learned with the actor network, where the inputs 
are the current and previous states (i.e., CSs) of each EV driver, actions, 
and rewards. The critic network uses a DNN to learn the action-value function 
$Q_w(s, a)$, which is used to update the parameters of the actor in the 
direction of reward improvement. The critic network is only needed for 
guiding the actor during training whereas only actor network is required at 
test stage. We update the parameter $w$ via minimizing 
$$J(w)=\mathbb{E}_{r_t,s_t \sim \rho^{\hat{\mu}(s)}} 
[(Q_w(s_t,a_t)-y^t)^2],$$ in which $y^t = r(s_t, a_t) + \gamma 
Q_w^{tar}(s_{t+1}, a_{t+1})$, $a_{t+1} \sim \pi_{\theta}(s_{t+1})$ is 
the charging action recommended by the actor network, and $\delta = 
(Q_w(s_t,a_t)-y^t)$ is Temporal Difference (TD) error, which is used for 
learning the Q-function. 

\subsection{The RAC Framework for EV Charging Recommendation}
In the previous formulation, we assume the state of an EV driver is fully 
observable. However, we are often unable to observe the full states of
an EV driver. Here we reformulate the environment of RAC as
Partially Observable Markov Decision Process (POMDP). In POMDP, $O$ is used to
denote the observation set, and we obtain each observation $o\in O$ directly 
from $p(o_t|s_t)$. For simplicity, we use a stacked LSTM together with the  
previous Fully Connected (FC) layers for each input step (Fig. 
\ref{fig:Neural_Networks}B), to summarize previous observations to substitute 
the partially observable state $s_t$ with $c_t = f_{\phi_1}(o_{t-k},..., 
o_{t-2}, o_{t-1})$. Each $o = (l, e,\tau)$ represents a observation in 
different time points, and $\phi_1$ is the set of parameters of $f$. $l$ 
denotes the CS location context information, $e$ presents charging event 
related features (e.g., SOC), and $\tau$ represents the time point (e.g., day 
of a week and hour of a day). $l$ is a combination context with the geodesic 
distance from previous CS (calculated by the latitudes and longitudes), one-hot 
encoding of this CS and the POI distribution around this CS. $o$ is a 
concatenation of $l$ , $e$ and $\tau$ vectors. The samples for training the 
actor model is generated from the behavior actor $\beta(s|a)$ (i.e., from the 
real world charging trajectories) via a buffer in an off-policy setting.   

Our RAC consists of three main DNN modules for estimating the actor, the 
critic, and the reward, as shown in Fig. \ref{fig:Neural_Networks}. Actor DNN 
(Fig. \ref{fig:Neural_Networks}B) captures each driver's charging 
preference. We take a subsequence of the most recent CSs as input to extract 
the hidden state $c_t$ through a stacked-LSTM. With the following fully 
connected layers, we recommend the CS to go next for an EV driver. During 
training, the actor is supervised with the TD from the critic network to 
maximize the expected reward and the actual CS selection from this driver with 
cross-entropy loss to minimize the difference (Fig. 
\ref{fig:Neural_Networks}A) . Since the actor is on each 
EV and takes private charging information as input, it is a 
driver-centered charging recommendation model.   

To enable resource-awareness, we use a one-way information 
transmission scheme, shown in Fig. \ref{fig:general}. We train a 
resource-aware actor for each EV driver via estimating $Q$ value from 
the critic DNN with addition of the immediate reward $r$ estimated with a 
reward DNN. Fig. \ref{fig:Neural_Networks}C shows the prediction of 
$Q_w$ value of state $s$ and action $a$, and the state here would be 
substituted with $c$ in POMDP setting. Fig. \ref{fig:Neural_Networks}D 
describes how to estimate the wait time $\hat{z}$ in all CSs. 
We can calculate the immediate reward for each pair of $(s, a)$ by combining 
with the estimated drive time. To tackle the cold start problem for new EV 
drivers, we introduce a warm-up training technique to update the model and 
will illustrate the details in the experiment section.

\subsection{Timely Estimation of Reward}


In stead of using traditional static reward, we dynamically estimate the 
reward from external environment. Since the drive time to and wait time at 
the CS play a key role for private EV driver's satisfaction, we estimate 
rewards based on these two factors. 
Specifically, we directly use the geodesic distance from map to represent the 
drive time and use a DNN (Fig. \ref{fig:Neural_Networks}D) to timely 
estimate the wait time for each charging. Therefore, a timely estimation of 
reward for choosing each CS can be given by a simple equation: $ \hat{r}_t = 
-100(\frac{\hat{z}_t}{\tilde{z}} + \zeta\frac{\hat{d}_t}{\tilde{d}})$ 
where $\hat{z}_t = g_{\phi_2}(l_t, \tau_t,z_{t-k}, \dotso, z_{t-1})$ is the 
predicted wait time through reward network, in which $\phi_2$ is the parameters 
of the reward DNN (LSTM).$z_{t-k}$ is the wait time in $k$ steps before the 
current time step, and it is directly summarized from the dataset we used. 
$\hat{d}_t$ is the estimated driving distance to the 
corresponding CS. Further, $\tilde{d}$ and $\tilde{z}$ 
represent statistically averaged driving distance and wait time in each CS, and 
they are constant values for a specific CS. $\zeta$ is a coefficient, which 
usually has an inverse relationship with an EV driver's familiarity with the 
routes (visiting frequency of each CS). For simplicity, we set $\zeta$ as 0.8 
for the most visited CS, and 1 for other situations. To make the predicted wait 
time and predicted driving distance to be additive, we do normalization for the 
predicted values by the averaged wait time $\hat{z}$ and $\hat{d}$ respectively 
for each CS. Since the wait time and drive time are estimated by each CS, our 
RAC framework is resource-aware to make CS recommendation for each EV driver.


Putting all the components as mentioned above together, the training algorithm 
of RAC is shown in Algorithm \ref{alg:training}.

\begin{algorithm}[h]
	\scriptsize
	\centering
	\caption{The RAC training algorithm}
	\label{alg:training}
	
	\begin{flushleft}
		\textbf{Input:} Actions $A$, observations $O$, reward function $r$, 
		$\#$ of CSs $M$, historical wait time at each hour $(z_1^j,...,z_T^j)$, 
		and coordinates $(\text{latitude}^j, \text{longitude}^j)$ in $j$-th  
		CS\\
		
		\textbf{Hyper-parameters}: Learning rate $\alpha = 0.001$, 
		$\epsilon=0.5$, 
		the finite-horizon step $T=10$, number of episodes $I$, and 
		$\gamma=0.99$\\
		
		\textbf{Output:}  $\theta, \phi_1, \phi_2, w$
	\end{flushleft}
	
	\begin{algorithmic}[1] 
		\STATE Store sequences $(o_1,a_1,r_1,...,o_T,a_T,r_T)$ by behavior 
		policy $\beta(a|s)$ in buffer $D$, each $o 
		= (l, e,\tau),$ and $\#$ of epochs $N$;\\
		\STATE{Random initialize actor $\pi_{\theta}$, critic $Q_w$, target 
			critic $Q_w^{tar}$, TD error $\delta=0$, and reward network 
			$f_\phi$;} \\
		\FOR{ $n=1$ to $N$ } 
		\STATE {Sample 
			$(o_1^i,a_1^i,r_1^i,...,o_T^i,a_T^i,r_T^i)\subset{D}$, 
			$i=1,...,I$} 
		
		\STATE{$c_t^i \leftarrow 
			f_{\phi_1}(o^i_{t-k},o^i_{t-k},...,o_{t-1}^i)$}
		\STATE ${a}_t^i, {a}_t^{i+1} \leftarrow$ sampled by $\pi_\theta$
		
		\STATE{$\hat{z}_t^i \leftarrow 
			g_{\phi_2}(l_t,\tau_t,z_{t-k},...,z_{t-1})$  }
		\STATE{$\hat{r}_t^i \leftarrow -100(\frac{\hat{z}_t^i}{\tilde{z}^i} 
			+ 
			\zeta\frac{\hat{d}_t^i}{\tilde{d}^i})$}
		\STATE{$y_t^i \leftarrow \hat{r}_t^i + \gamma 
			Q_w^{tar}(c_t^{i+1},{a}_t^{i+1}) $}
		\STATE $\hat{a}_t^i \leftarrow$ given by the EV driver's selection
		\STATE $\delta^i_t \leftarrow Q_w(c_T^i,\hat{a}^i_t)-y_t^i$
		\STATE $w \leftarrow 
		w-\alpha\frac{1}{IT}\sum_i\sum_t\delta^i_t \nabla_wQ_w(c_t^i,a_t^i)$
		\STATE $\phi_1 \leftarrow \phi_1 
		-\alpha\frac{1}{IT}\sum_i\sum_t\delta^i_t \nabla_{\phi_1} 
		f_{\phi_1}$
		\STATE $\phi_2 \leftarrow \phi_2 
		-\alpha\frac{1}{IT}\sum_i\sum_t\delta^i_t \nabla_{\phi_2} 
		g_{\phi_2} $
		
		\STATE $\nabla_wQ_w(c_t^i,a_t^i) \leftarrow$ given by 
		$Q_w(c_t^i,a_t^i)$
		\STATE $\eta_t^i=\frac{1}{M}\sum_{k=1}^{M} 
		\frac{\hat{a}_{t,k}^i-a_{t,k}^i}{(1-a_{t,k}^i)a_{t,k}^i}$
		\STATE $\theta \leftarrow \theta + 
		\alpha\frac{1}{IT}\sum_i\sum_t[(1-\epsilon)\nabla_wQ_w(c_t^i,a_t^i)+\epsilon\eta_t^i]$
		\ENDFOR
	\end{algorithmic}
\end{algorithm}

\subsection{Geospatial Feature Learning}\label{sec:geospatial}
The POI distribution within the neighborhood of each CS is what we used to 
learn the geospatial features from each CS. With this information, we can infer 
the semantic relationships among the CSs to assist in recommending CSs for each 
driver. Google Map defines 76 types (e.g., schools, restaurants, and hospitals) 
of POIs. Specifically, for each CS, we use its latitude and longitude 
information together with a geodesic radius of 600 meters to pull the 
surrounding POIs. We count the number of POIs for each type to obtain a 
76-dimension vector (e.g., $POI \in \mathbb{R}^{76}$) as the POI distribution. 
We concatenate this vector with other information, i.e., geodesic distances to 
CSs and one-hot encoding of the CS. With the charging event features and the 
timestamp-related features, we learn a unified embedding through an MLP for 
each input step of the stacked-LSTM.

\section{Experiments}
\subsection{Experimental Setup}
All the experiments are implemented on two real world charging 
events datasets from Dundee city 
\footnote{https://data.dundeecity.gov.uk/dataset/} and Glasgow city 
\footnote{http://ubdc.gla.ac.uk/dataset/}. The POI distribution for each CS is 
obtained from Google Place 
API\footnote{https://developers.google.com/maps/documentation/places/web-service}.
 The code of our method is publicly available on this link: 
 https://github.com/cyli2019/RAC-for-EV-Charging-Rec.

\subsubsection{Datasets and Limitations}
For Dundee city, we select the charging events from the time range of 
6/6/2018-9/6/2018, in which there are 800 unique EV drivers, 44 CSs and 19, 115 
charging events. For Glasgow city, in the time range of 9/1/2013-2/14/2014, we 
have 47 unique EV drivers, 8 CSs and 507 charging events. For each charging 
event, the following variables are available: CS ID, charging event ID, EV 
charging date, time, and duration, user ID, and consumed energy (in kWh) for 
each transaction. For each user ID, we observe a sequence of charging events in 
chronological order to obtain the observations $O$. For each CS ID, we learn 
the geospatial feature to determine their semantic similarity according to POI 
types.  

To model an EV driver preference, we train a model using the CS at each 
time point as the outcome and the previous charging event sequence as the 
input. To enable situation awareness, for a specific CS, there is a chronically 
ordered sequence of wait time, and we use the wait time corresponding to each 
time point as the outcome and that of previous time points as inputs in our 
reward network to forecast hourly wait time for all CS's. Combined with the 
estimated drive time that are inverse proportional to familiarity adjusted 
geodesic distance, we determine the timely reward for each EV driver's charging 
event.

To our knowledge, these two datasets are the only publicly 
available driver-level charging event data for our driver-centered charging 
recommendation task, though with relatively small size and unavailability of 
certain information. Due to the privacy constraints, the global positioning 
system (GPS) information of each driver and the corresponding timestamp are not 
publicly available as well as traffic information in these two cities during 
the time frame. As such, we have no choice but having to assume the EV driver 
transits from CS to CS and using driving distance between CSs combined with 
estimated wait time at each CS to calculate the external reward. Another
assumption we made is using the time interval of each charging event to 
approximate the SOC of the EV since all EVs in the data sets are of the same 
model. The method developed in this paper is general that does 
not rely on the aforementioned assumption; when GPS, timestamp and SOC 
information become available, our method is ready to work without change.

\subsubsection{Evaluation Metrics} Similar to POI recommendation, we 
treat the earlier 80\% sequences of each driver as a training set, the middle 
10\% as a validation set, and the latter 10\% as a test set. Two standard 
metrics are adopted to evaluate methods' performance, namely, Precision (P@K) 
and Recall (R@K) on the test set. To quantify the external reward for 
making a charging recommendation, we also use a Mean Average Reward (MAR) as an 
evaluation metric. Each reward is calculated based on familiarity-adjusted 
geodesic distance and projected wait time at the recommended CS, and MAR is the 
average value over all users across all time points in the test set. To solve 
the cold-start problem for EV drivers who have few charging events, we use 5\% 
of data in the earlier sequences from all users (with more than 10 charging 
events) to train a model as warm-up, the rest 95\% following the same data 
splitting strategy described above followed by training with each driver's 
private data. We assume that for the earliest 5\% of data can be shared without 
privacy issues when the user related information is eliminated.  

\begin{table}
	\caption{Performance comparison with different learning methods. Results of 
		the best-performing RAC model are boldfaced; the runner-up is 
		labeled with '*'; `Improvement' refers to the percentage of improvement 
		that RAC achieves relative to the runner-up results. }
	\scriptsize
	\label{tab:inter_comparasion}
	\begin{tabular}{| c | c c c c c c c c c|}
		\hline
		Dataset & Metrics & MC & FPMC & Time-LSTM &ST-RNN & 
		ATST-LSTM  & RAC-zero & RAC & Improvement \\
		\hline
		\multirow{7}{*}{Dundee}  
		&$P$@1 & 0.204 & 0.242 & 0.313 & 0.326 & 0.368* & 0.385 
		& 
		\textbf{0.424} & 15.2\%\\
		&$P$@3 & 0.256 & 0.321  & 0.367 & 0.402 & 0.435* & 0.463 
		& \textbf{0.509} & 17.0\%\\
		&$P$@5 & 0.321 & 0.363  & 0.436 & 0.437 & 0.484* & 0.528 
		& \textbf{0.577}  & 19.2\%\\
		&$R$@1 & 0.146 & 0.195  & 0.203 & 0.216 & 0.247* & 
		0.285  & \textbf{0.292} & 18.2\% 	\\
		&$R$@3 & 0.153 & 0.226 & 0.236 & 0.278  & 0.298* & 
		0.344 & \textbf{0.368}  & 23.5\%	\\
		&$R$@5 & 0.192 & 0.237  & 0.245  & 0.325 & 0.375* & 
		0.427 & \textbf{0.479}  & 27.7\%\\
		&$MAR$ & -327.8 & -265.9 & -210.4 & -195.4 & -164.5* & 
		-133.2 & \textbf{-114.6}  & 30.3\% 
		\\
		\hline
		\hline
		\multirow{7}{*}{Glasgow}  
		&$P$@1 & 0.163 & 0.207 & 0.264 & 0.252 & 0.294* & 0.313 
		&\textbf{0.364}  & 23.8\%				\\
		&$P$@3 & 0.226 & 0.262  & 0.325 & 0.356 & 0.375* & 
		0.40.9 & \textbf{0.458}  & 22.1\%\\
		&$P$@5 & 0.285 & 0.301 & 0.398 & 0.405 & 0.428* & 0.482 
		& 
		\textbf{0.497} & 16.1\%\\
		&$R$@1 & 0.108 & 0.093  & 0.122 & 0.128 & 0.133* & 
		0.13.1 & 
		\textbf{0.164}  & 23.3\%\\
		&$R$@3 & 0.126 & 0.135  & 0.174 & 0.182 & 0.216* & 
		0.224 & 
		\textbf{0.253}  & 17.1\%\\
		&$R$@5 & 0.173 & 0.167  & 0.263 & 0.323 & 0.334* & 
		0.395 & 
		\textbf{0.406}  & 21.5\%\\
		&$MAR$ & -456.3 & -305.4  & -232.2  & -210.9  & -196.4* 
		& -164.2 & \textbf{-154.3}  & 21.4\%\\
		\hline
	\end{tabular}
\end{table}

\subsubsection{Baselines}  We compare RAC with the following baseline 
methods, including two classic methods (i.e., MC, and FPMC 
\cite{rendle2010factorizing}), three DNN-based state-of-the-art methods (i.e., 
Time-LSTM \cite{zhu2017next} , ST-RNN \cite{liu2016predicting}, and ATST-LSTM 
\cite{huang2019attention}). We select these methods as the baselines for method 
comparison, instead of other general POI recommendation methods (e.g., 
multi-step or sequential POI recommendation problem), because they directly 
address the next POI recommendation problem. One variant of our RAC (i.e., 
RAC-zero) is trained from scratch without warm-up training. The description of 
the baselines are: (1) MC: first-order Markov Chain utilizes 
sequential data to predict a driver's next action based on the last actions via 
learning a transition matrix. (2) FPMC: Matrix factorization method learns the 
general taste of a driver by factorizing the matrix over observed driver-item 
preferences. Factorization Personalized Markov Chains model is a combination of 
MC and MF approaches for the next-basket recommendation.  (3) Time-LSTM: 
Time-LSTM is a state-of-the-art variant of LSTM model used in recommender 
systems. Time-LSTM improves the modeling of sequential patterns by explicitly 
capturing the multiple time structures in the check-in sequence. We used the 
best-performing version reported in their paper. (4) ST-RNN : It is a RNN-based 
method that incorporates spatiotemporal contexts for next location prediction. 
(5) ATST-LSTM: It utilizes POIs and spatiotemporal contexts in a 
multi-modal manner for next POI prediction. In addition, to evaluate the effect 
of warm-up training on solving the cold-start problem, we compare our RAC with 
its a variant, RAC-zero, which is trained from scratch.


\subsection{Performance Comparison}
The parameter tuning information during the training are described above, and 
after that we make comparison for our approach with the baselines methods.Table 
\ref{tab:inter_comparasion} presents the performance (R@K, P@K, and MAR) of all 
methods across the two datasets. We test $K$ with 1, 3, and 5, and based on the 
parameter tuning results, we use the setting of two-layer stacked-LSTM for both 
actor and reward networks, embedding/hidden sizes of (100, 100), $\epsilon$ of 
0.5, and learning rate of 0.001. The feeding steps for LSTMs in actor and 
reward networks are set to 5 and 10 respectively. In terms of charging 
recommendation task, the RNN based methods (Time-LSTM, ST-RNN, ATST-LSTM, and 
RAC) generally outperforms non-RNN based competitors (MC, and FPMC) owing to 
the leverage of spatiotemporal features. For the former, ATST-LSTM is better 
than ST-RNN possibly due to the effective use of attention mechanism. ST-RNN 
has slightly better performance over Time-LSTM due to the incorporation of 
spatial features. Overall, our proposed RAC consistently achieves the best 
performance not only on precision/recall but also over MAR, in which the 
improvement column are the comparisons between RAC and the runner-up model 
(ATST-LSTM). This is translated into the fact that overall RAC is capable of 
accommodating inherent user preference and ensuring the external rewards to a 
maximum extent in rendering charging recommendations. 

\begin{figure}[t]
	\centering
	\scriptsize
	\includegraphics[scale=0.6]{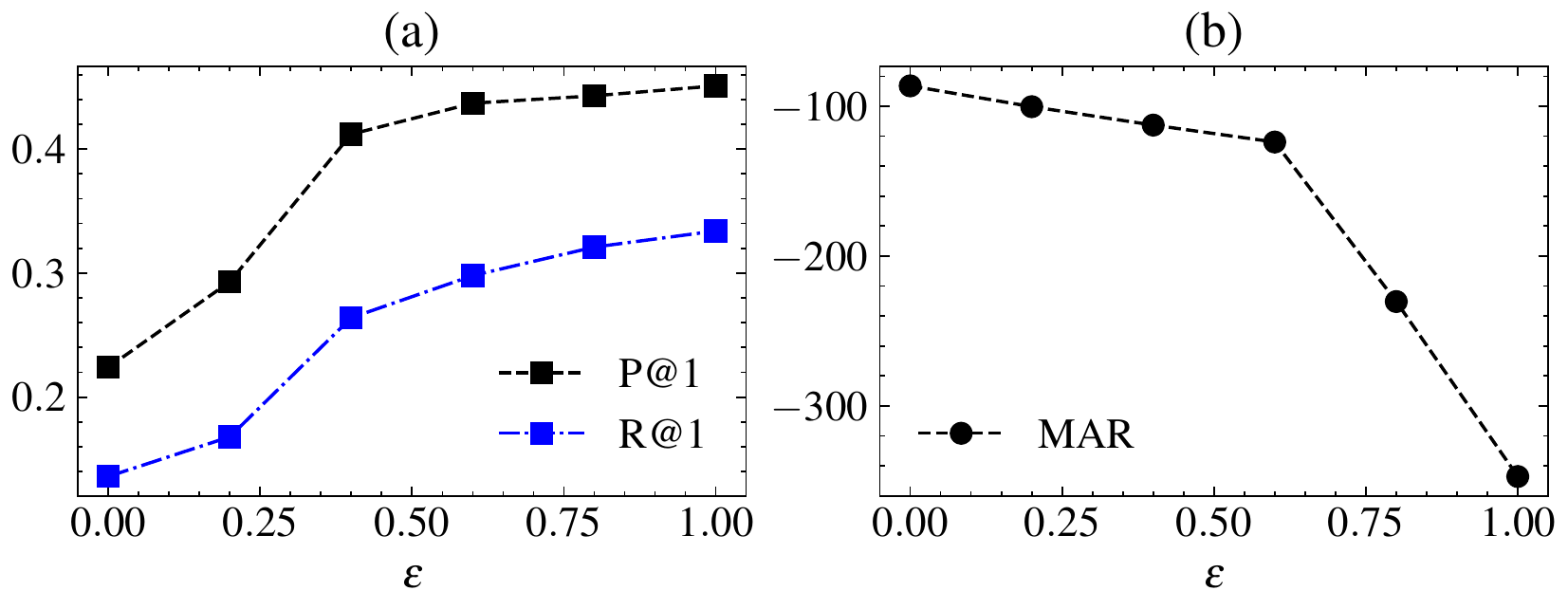}
	\caption{The tension between maximizing (a) inherent user 
		preference and (b) external reward on the averaged user.}
	\label{fig:parameter_sensitivity}
\end{figure}

\begin{figure}[t]
	\centering
	\scriptsize
	\includegraphics[scale=0.6]{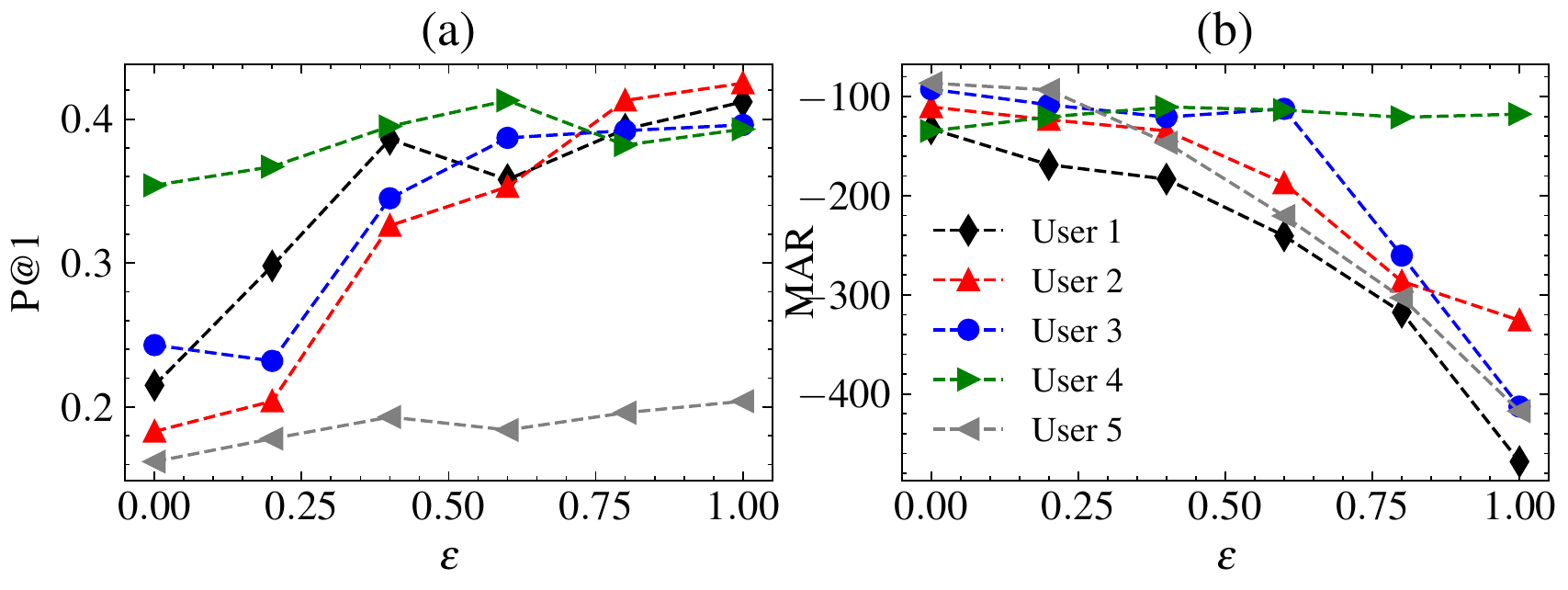}
	\caption{A case study of five individual EV drivers (a) inherent user 
		preference, and (b) external reward.}
	\label{fig:specific_user}
\end{figure}

To demonstrate the influence of warm-up training in RAC, we compare it with the 
training-from-scratch-approach RAC-zero. From Table \ref{tab:inter_comparasion},
RAC demonstrates a better overall performance on the Dundee dataset for the 
relative abundance of samples for warm-up training; in the meanwhile, due to 
the limited number of warm-up training samples in the Glasgow dataset, this 
improvement is relatively slight. Conventionally, the Glasgow dataset with 
fewer charging stations might have better recommendation accuracy than the 
Dundee dataset. However, we should know that most (over 80\%) EVs are 
revisiting no more than eight charging stations for both datasets. Therefore, 
for driver-centered charging pattern, the number of possible CSs is similar 
for these two datasets, resulting in even worse performance for the Glasgow 
dataset than the Dundee dataset. Overall, our proposed RAC consistently 
achieves the best performance not only on precision/recall but also over MAR.

\subsection{Driver-centered CS Recommendation}

Fig. \ref{fig:parameter_sensitivity} illustrates the effect of 
personalization tuning parameter $\epsilon$ on precision/recall and reward of 
the recommendation. Since RAC is a driver-centered recommendation method, each 
driver can experiment with the parameter $\epsilon$ to weigh more on inherent 
user preference or on external award when seeking driver-centered charging 
recommendations. In Fig. \ref{fig:parameter_sensitivity}A, the P@1 and R@1
of RAC climb up as $\epsilon$ increases, and becomes stable at around 0.5, 
indicating a larger value would not further improve the performance. In Fig.
\ref{fig:parameter_sensitivity}(b), MAR first decreases slightly before 0.5 and 
then drops quickly afterwards. Collectively, it appears an average driver can 
get the best of both worlds when $\epsilon$ is around 0.5. 

Fig. \ref{fig:specific_user} shows that drivers 1-3 follow a very similar 
pattern to the average driver in Fig. \ref{fig:parameter_sensitivity} where 
$\epsilon$ is around 0.5, representing a good trade-off to balance between the 
inherent user preference and the external reward. Driver 4 represents a special 
case where the driver preference aligns well with the external reward; in this 
case the charging recommendation is invariant to the choice of $\epsilon$. 
Hence the recommendation can be made either based on user preference or 
external reward since they are consistent to each other. Driver 5 represents a 
new driver with low precision and recall due to the lack of historical charging 
data. As such, the recommendation can simply be made based mostly on the 
external reward via setting $\epsilon$ to a low value, e.g., 0.2. In sum, 
tuning $\epsilon$ indeed enables an individual driver to be more attentive to 
his/her preference or to the external reward when seeking EV charging 
recommendation. 

\begin{figure}[t]
	\centering
	\scriptsize
	\includegraphics[scale=0.42]{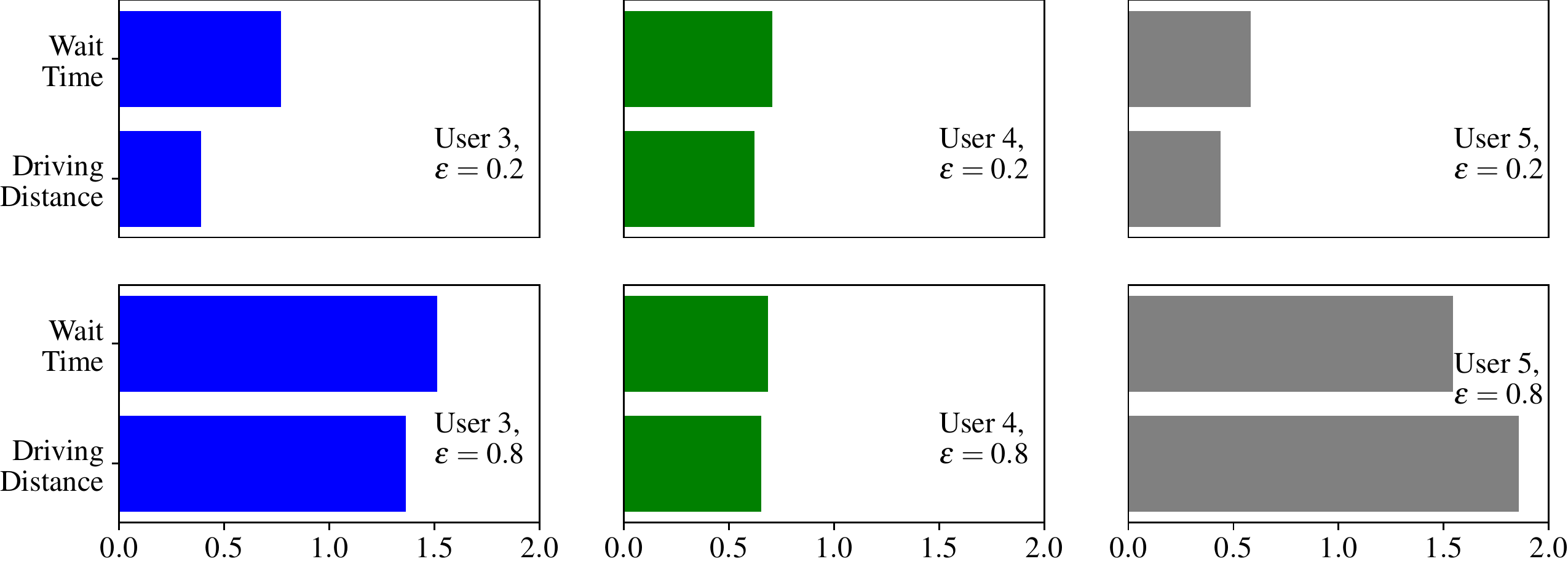}
	\caption{Examples of explanations for the recommendations. We use the mean 
		value of the normalized wait time and driving distance to make the 
		comparison fair.}
	\label{fig:explain}
\end{figure}

Fig. \ref{fig:explain} demonstrates the award (e.g., wait time and driving 
distance) for three representative EV drivers, User 3, User 4, and User 5, 
under two different values of $\epsilon$, i.e., 0.2 and 0.8. Recall the latter 
denotes the weight on an EV driver to follow historical  
charging pattern. Therefore, an increase of the $\epsilon$ value from 
0.2 to 0.8 indicates that the charging recommendation is rendered based more on 
the driver's previous charging pattern than the reward from external 
environment. In Fig. \ref{fig:explain}, we describe three types of drivers 
demonstrated by different trade-offs: (1) For User 3, the wait time and driving 
distance are both increasing, resulting in a smaller reward, whereas a 
better prediction accuracy. (2) For User 4, the wait time and driving distance 
remains shorter yet stable across the two values of $\epsilon$, demonstrating 
both a larger reward and higher prediction accuracy. (3) For User 5 who is a 
newer driver, the reward increases similarly to User 3. However, the prediction 
accuracy stays low regardless of the choice of $\epsilon$ due to the limited 
information on historical charging pattern of the new driver. 
In summary, for drivers such as User 3 whose charging patterns are vastly 
deviated from what would be recommended by the external award, tuning 
$\epsilon$ would allow the drivers to be more attentive to either historical 
charging patterns or the external award. For drivers such as User 4 whose 
historical charging pattern is consistent with the more rewarding charging 
option as determined by shorter wait time and driving distance, the choice of 
$\epsilon$ does not matter, representing an optimal charging recommendation 
scenario. For new drivers such as User 5, a charging recommendation that is 
largely based on the external reward may be more appropriate.     

\section{Conclusion}
In this paper, we propose a resource-aware and driver-centered charging 
recommendation method for private EVs. We devise a flexible regularized 
actor-critic framework, i.e., using RL to maximize external reward as the 
regularization to model inherent user preference for each driver. Our approach 
is sufficiently flexible for a wide range of EV drivers including new drivers 
with limited charging pattern data. Experimental results on real-world datasets 
demonstrate the superior performance of our approach over the state-of-the-arts 
in the driver-centered EV charging recommendation task.

\bigskip
\noindent{\bf Acknowledgements}  
This work is supported by the National Science Foundation under grant no. IIS-1724227.

%
%
%
%
\bibliographystyle{splncs04}
\bibliography{refs}
\end{document}